%% file: main.tex
\documentclass[10pt,twocolumn,letterpaper]{article}

\usepackage[pagenumbers]{cvpr} % To force page numbers, e.g. for an arXiv version
\usepackage{adjustbox,booktabs,graphicx,multirow, amssymb, graphicx}

\input{preamble}
\definecolor{cvprblue}{rgb}{0.21,0.49,0.74}
\usepackage[pagebackref,breaklinks,colorlinks,allcolors=cvprblue]{hyperref}

\def\confName{CVPR}
\def\confYear{2027}

\title{Learning and Comparing Local Traces for Weakly Supervised Image Manipulation Localization}

\author{
Xinghao Wang \and
Changtao Miao \and
Dianmo Sheng \and
Tao Gong \and
Qi Chu \and
Nenghai Yu \and
Quanchen Zou \and
Deyue Zhang \and
Xiangzheng Zhang
}

\begin{document}
\maketitle
\input{sec/0_abstract}    
\input{sec/1_intro}
\input{sec/2_related}
\input{sec/3_method}
\input{sec/4_experiment}
\input{sec/5_conclusion}
\input{sec/6_LLM}
{
    \small
    \bibliographystyle{ieeenat_fullname}
    \bibliography{main}
}

% WARNING: do not forget to delete the supplementary pages from your submission 
% \input{sec/X_suppl}

\end{document}

%% file: sec/0_abstract.tex
\begin{abstract}
Image manipulation localization identifies edited regions
and provides spatial evidence for assessing image reliability,
but fully supervised methods require costly pixel-level
annotations. Weakly supervised methods reduce this cost
using image-level labels, but classification can rely on
a few salient cues without ensuring accurate or complete
localization. This difficulty is pronounced when manipulation
traces are weak and unevenly distributed and authentic
and manipulated regions share similar appearance.
We propose a CLIP-based framework that integrates
Trace-Oriented Vision--Language Adaptation (TOVA)
and Relation-Aware Trace Discrimination (RATD).
TOVA adapts the visual encoder and uses global visual
context to aggregate edge-weighted patch features into
image-conditioned class prompts. The resulting class
representations support image-level alignment and guide
patch selection for Hierarchical Trace Contrastive Learning,
providing local contrastive supervision across visual layers.
RATD combines local context, spatial structure, and
original-image details, then compares learned descriptors
between appearance-matched regions. Bidirectional comparison
messages are integrated with local features for dense
localization. The two components share a visual encoder
and are jointly trained using only image-level authenticity
labels. Experiments on multiple public benchmarks demonstrate
that our framework outperforms the evaluated weakly supervised
baselines in image manipulation localization.
\end{abstract}

%% file: sec/1_intro.tex
\section{Introduction}
\label{sec:introduction}

Image editing tools can modify visual content while
preserving a plausible appearance. When manipulated
images are used to misrepresent events or evidence,
identifying the edited regions helps assess their
reliability. Image manipulation localization provides
this spatial evidence by predicting the location and
extent of modifications.

Fully supervised methods learn localization from forensic
traces using pixel-level annotations.
CAT-Net models compression artifacts in the RGB and
DCT domains~\cite{kwon2021cat}, while MVSS-Net combines
noise and boundary features with pixel-, edge-, and
image-level supervision~\cite{chen2021image}.
Obtaining pixel-level annotations for large and diverse
image collections is costly, motivating weakly supervised
localization using only image-level authenticity
labels~\cite{zhai2023towards}.

Recent methods derive regional learning signals through
multi-source and inter-patch consistency~\cite{zhai2023towards},
tampering-edge activation maps~\cite{zhou2024exploring},
and vision--language reasoning with contrastive patch
consistency~\cite{sheng2026weakly}.
Under image-level supervision, however, correct
classification does not ensure accurate or complete
localization. A model can rely on a few salient cues
without identifying the full manipulated region,
particularly when manipulation traces are weak and
unevenly distributed.

\begin{figure}[t]
\centering
\includegraphics[width=0.48\textwidth]{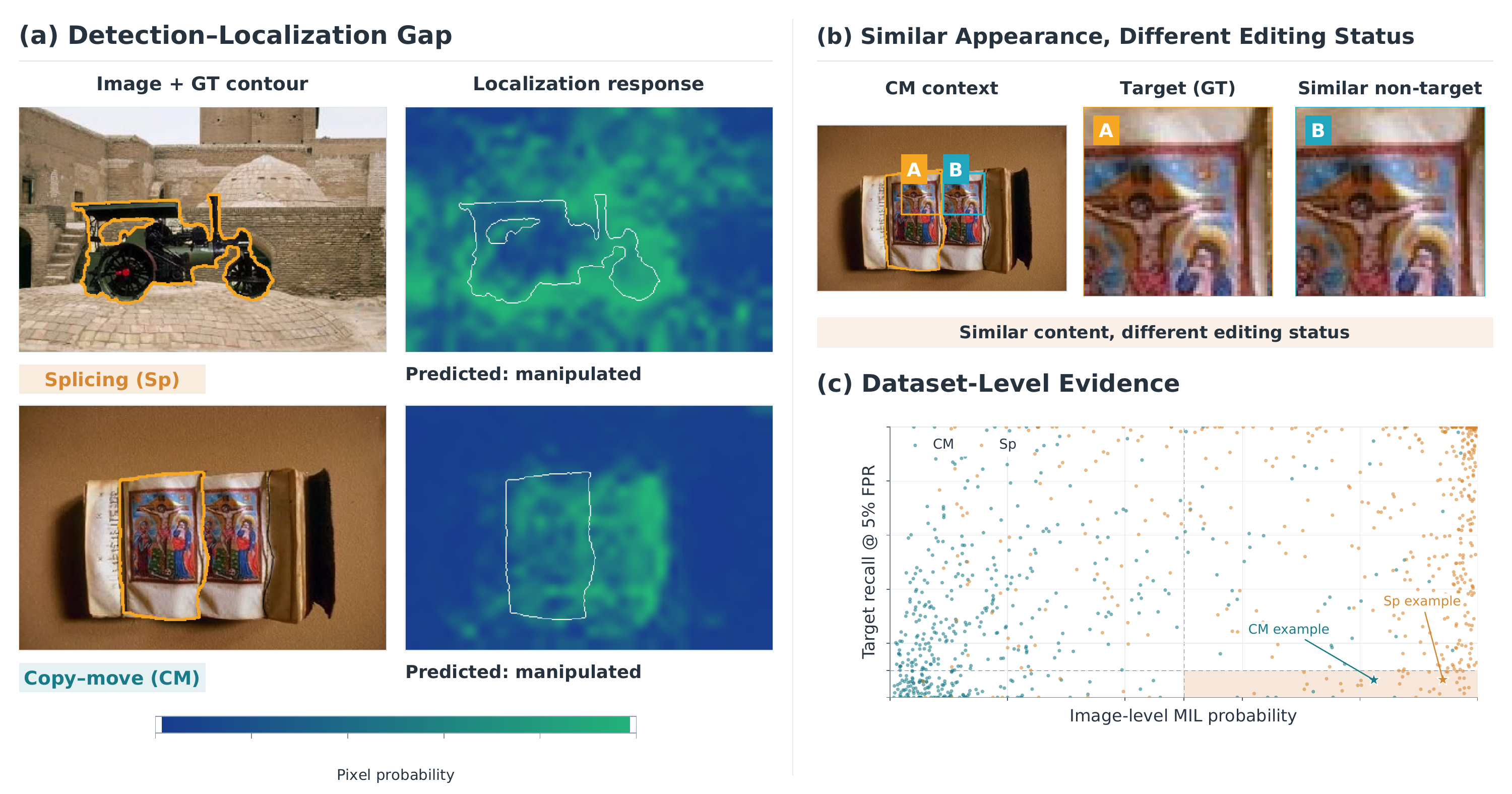}
\caption{Motivation for local trace learning and regional
comparison, illustrated using a fine-tuned CLIP model on
CASIAv1.
(a) The splicing (Sp) and copy-move (CM) examples are correctly
detected as manipulated with image-level scores of 0.941
and 0.824, respectively, but their target recall is only
6.6\% and 6.5\%. Heatmaps show pixel-level probabilities
on a shared $[0,1]$ scale.
(b) Target and non-target regions from the same image
have similar appearance despite different manipulation
labels. Both crops contain $64\times64$ pixels and are
shown at the same magnification.
(c) Dataset-level analysis covers 920 manipulated images
(459 CM and 461 Sp). Among 414 correctly detected images,
40 (9.7\%) have target recall below 10\%, including
7/49 CM images (14.3\%) and 33/365 Sp images (9.0\%).
Target recall is measured offline using ground-truth masks
under a background false-positive rate of at most 5\%.
These observations show that correct image-level decisions
can coexist with poor regional localization, while similar
appearance does not determine local authenticity.
They motivate learning local manipulation representations
and interpreting them through regional comparisons.}
\label{fig:moti}
\vspace{-0.3cm}  % 减少表格和正文之间的空白
\end{figure}

Moreover, local features encode both image content
and processing traces. Authentic and manipulated regions
may share similar appearance, making their authenticity
difficult to determine from appearance alone.
Boundary cues illustrate this ambiguity: discontinuities
can indicate editing traces~\cite{chen2021image,zhou2024exploring},
but similar responses may also occur at natural edges.
These difficulties motivate learning representations
sensitive to local manipulation traces and interpreting
their evidence relative to surrounding content and
appearance-matched regions.

CLIP provides transferable visual and textual
representations~\cite{radford2021learning} that can be
adapted through learnable prompts~\cite{zhou2022learning}
and image-conditioned prompting~\cite{zhou2022conditional}.
We propose a CLIP-based framework that connects image-level
authenticity learning with local trace supervision and
regional evidence comparison. It consists of
\emph{Trace-Oriented Vision--Language Adaptation} (TOVA)
and \emph{Relation-Aware Trace Discrimination} (RATD).
The two components share a visual encoder and are jointly
trained using only image-level authenticity labels.

TOVA adapts the visual encoder and learns image-conditioned
class prompts. Global visual context guides the aggregation
of edge-weighted patch features to construct prompts for
the authentic and manipulated classes.
The resulting class representations support image-level
alignment and guide patch selection for
\emph{Hierarchical Trace Contrastive Learning} (HTCL),
which applies contrastive supervision to selected patch
features across visual layers.

RATD compares local evidence using spatial context and
intra-image references. It combines contextual variation,
spatial structure, and original-image details into a
common representation, then compares learned descriptors
between appearance-matched regions.
Bidirectional comparison messages are integrated with
local features to predict manipulated regions.
Multiple instance learning supervises these spatial
predictions using image-level labels.

The main contributions are summarized as follows:
\begin{itemize}
    \item We introduce Trace-Oriented Vision--Language
    Adaptation, which combines prompts conditioned on
    local structural cues with hierarchical contrastive
    supervision to connect image-level authenticity
    learning with local manipulation representations.

    \item We develop Relation-Aware Trace Discrimination,
    which integrates local context and image details
    with bidirectional evidence comparisons between
    appearance-matched regions for dense localization.

    \item Experiments on standard image manipulation
    benchmarks demonstrate the effectiveness of the
    proposed framework under image-level supervision.
\end{itemize}

\begin{figure*}[t]
    \centering
    \captionsetup{font=small}
    \setlength{\belowcaptionskip}{-0.1cm}   %调整图片标题与下文距离
    \includegraphics[width=0.95\textwidth]{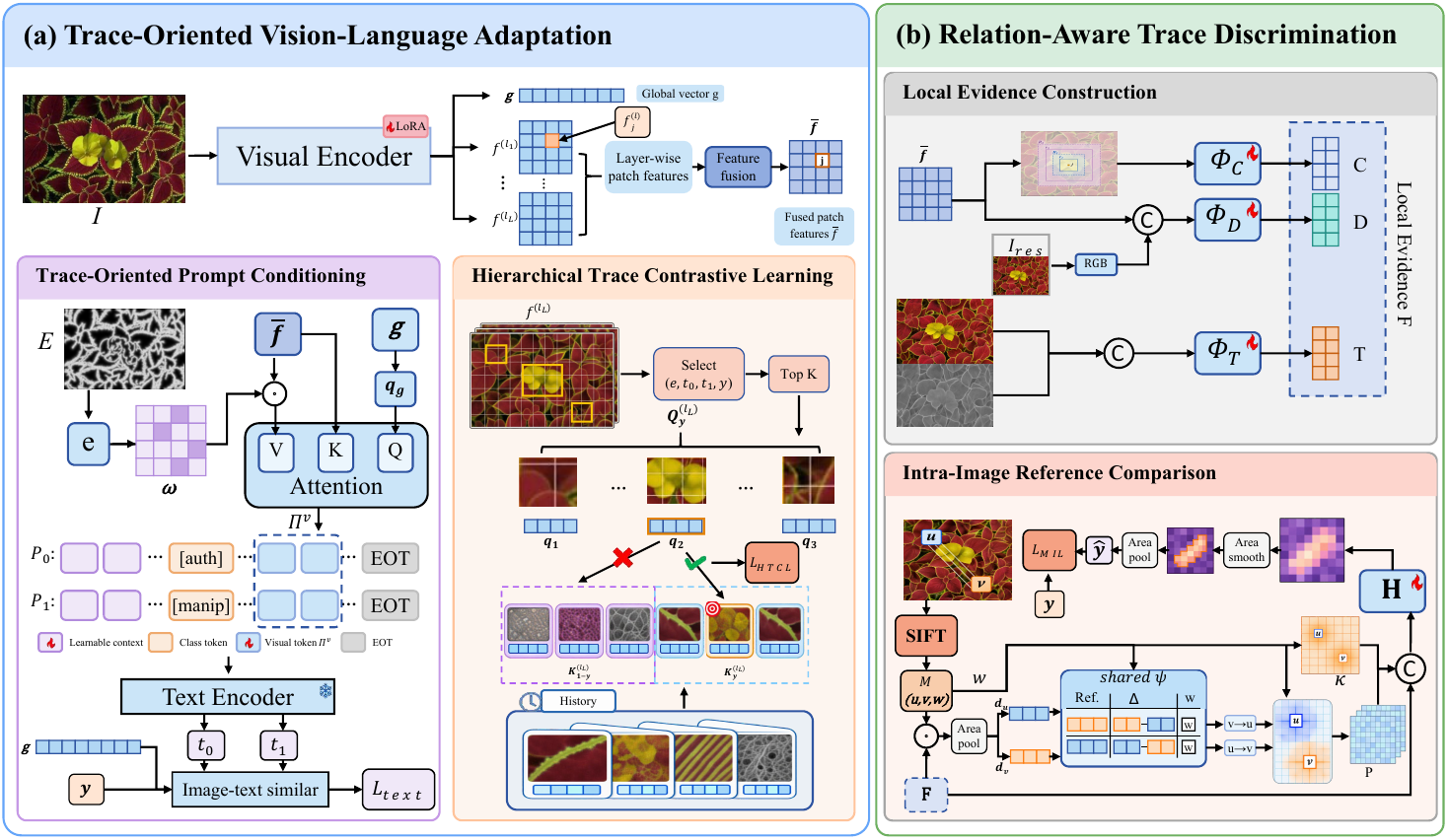}
    \caption{Overview of our framework for weakly supervised image
manipulation localization.
(a) Trace-Oriented Vision--Language Adaptation (TOVA) adapts
the CLIP visual encoder with LoRA and conditions authenticity
prompts on global visual context and edge-weighted patch
features. The resulting class representations guide patch
selection for Hierarchical Trace Contrastive Learning (HTCL)
across visual layers.
(b) Relation-Aware Trace Discrimination (RATD) integrates
contextual, structural, and original-image detail features
with bidirectional comparison messages between SIFT-matched
regions to predict manipulation maps.
Localization is supervised through area-weighted multiple
instance learning, and the framework is jointly trained
using only image-level authenticity labels.}
    \label{fig:framework}
    \vspace{-0.5cm}  % 减少表格和正文之间的空白
\end{figure*}

%% file: sec/2_related.tex
\section{Related Work}
\label{sec:related_work}

\subsection{Weakly Supervised Image Manipulation Localization}
\label{sec:related_weakly_supervised}

Early forensic methods exploit color filter array
artifacts~\cite{ferrara2012image} and local noise
statistics~\cite{mahdian2009using} without manipulation
masks, but rely on cue-specific assumptions.
Fully supervised methods use pixel-level annotations:
CAT-Net~\cite{kwon2021cat} models RGB/DCT compression
artifacts, MVSS-Net~\cite{chen2021image} combines noise
and boundary features with multi-scale supervision,
and SparseViT~\cite{su2025can} uses sparse self-attention
without handcrafted feature extractors.

Under image-level supervision, WSCL~\cite{zhai2023towards}
combines multi-source and inter-patch consistency with
online pseudo-label refinement.
M$^2$RL-Net~\cite{li2025m2rl} couples patch-level
self-consistency with feature-level contrastive learning
within and across RGB and noise views.
SO-WSL~\cite{zhu2025self} iteratively refines pseudo-labels
generated by the Segment Anything Model under class
activation map guidance.
WSCCL~\cite{bai2025weakly} uses visual and noise features
for cross-contrastive learning and adaptively fuses their
prediction maps into pseudo-labels.
EdgeCAM~\cite{zhou2024exploring} combines tampering-edge
responses from different training stages with
segmentation-based object proposals.
Our RATD combines local context with comparisons of learned
local evidence between appearance-matched regions within
an image.

\subsection{Vision-Language Models and Prompt Learning}
\label{sec:related_prompt_learning}

CLIP~\cite{radford2021learning} learns transferable visual
and textual representations through image--text contrastive
pretraining.
CoOp~\cite{zhou2022learning} learns continuous textual
context vectors, CoCoOp~\cite{zhou2022conditional} conditions
prompts on individual images, and
DenseCLIP~\cite{rao2022denseclip} combines pixel--text
matching with image-conditioned prompting for dense
prediction.

For manipulation localization,
Prompt-IML~\cite{liu2024multi} uses separate visual prompts
in a frozen backbone to process spatial and high-frequency
features, followed by feature alignment and fusion.
CLIP-IFDL~\cite{li2024noise} combines instance-aware
textual prompts with a noise adapter.
Both use pixel-level supervision.
Under image-level supervision,
ViLaCo~\cite{sheng2026weakly} combines learnable prompts,
adaptive vision--language interaction, dual prediction
heads, and contrastive patch consistency.

Our TOVA combines visual adaptation with prompts conditioned
on edge-weighted visual features and uses the resulting
class representations to select patches for contrastive
supervision across visual layers.

%% file: sec/3_method.tex
\section{Methodology}
\label{sec:method}

\subsection{Overview}
\label{sec:method_overview}

We study weakly supervised image manipulation localization using
only image-level authenticity labels. Image-level supervision
allows a model to classify an image based on a few salient cues,
but does not directly constrain the location or extent of
manipulated regions. This gap makes accurate localization
difficult when manipulation traces are weak and unevenly
distributed, especially when authentic and manipulated regions
share similar content and appearance.

As illustrated in Fig.~\ref{fig:framework}, we propose a framework
that adapts vision--language representations to local manipulation
traces and uses regional comparisons to guide localization.
The framework consists of two complementary components.

\emph{Trace-Oriented Vision--Language Adaptation} combines
visual adaptation and image-conditioned prompt learning in
CLIP~\cite{radford2021learning}, together with local contrastive
supervision, to learn representations sensitive to local
manipulation traces.
To further distinguish authentic and manipulated regions with
similar appearance, \emph{Relation-Aware Trace Discrimination}
uses local context and intra-image correspondences to identify
subtle processing differences and guide localization.
The following sections describe these two components and their
joint optimization.

\subsection{Trace-Oriented Vision--Language Adaptation}
\label{sec:trace_adaptation}

To strengthen local manipulation representations under
image-level supervision, we combine trace-oriented prompt
conditioning with hierarchical trace contrastive learning
(HTCL). Prompt conditioning incorporates local structural
cues into image--text learning. HTCL then uses the resulting
class representations to select patches for direct contrastive
supervision across visual layers. We adapt the shared CLIP
visual encoder using LoRA~\cite{hu2021lora}.

\paragraph{Trace-oriented prompt conditioning.}
Manipulation can alter local structural transitions while
preserving the overall image content. We therefore condition
class prompts on both visual content and edge proximity.
For an input image, we construct a continuous soft edge field:
\begin{equation}
E(u)
=
\left[
\mathcal{G}_{\sigma} *
\exp\!\left(-a_e\delta^{b_e}d_{\mathrm{edge}}\right)
\right](u)
\label{eq:method_soft_edge}
\end{equation}
where $d_{\mathrm{edge}}(u)$ is the distance from pixel $u$
to the nearest Canny edge pixel, $\delta$ is the fraction
of edge pixels in the image, and $\mathcal{G}_{\sigma}$
is a Gaussian smoothing kernel. The parameters $a_e$ and
$b_e$ control the decay as a function of edge density.

We map $E$ onto the visual patch grid and normalize its
responses across spatial locations. The resulting weights
condition the value input of attention, which aggregates
patch features using a global image query:
\begin{equation}
\begin{aligned}
\boldsymbol{\omega}
&= \operatorname{softmax}\!\left(\mathcal{T}_p(E)\right),\\
\Pi^v
&= \operatorname{MHA}\!\left(
q_g,\phi_k(\bar f),\phi_v(\boldsymbol{\omega}\odot\bar f)
\right).
\end{aligned}
\label{eq:method_edge_condition}
\end{equation}
Here, $\mathcal{T}_p$ maps the soft edge field to the visual
patch grid, with $e_j=[\mathcal{T}_p(E)]_j$ denoting the
edge response at patch location $j$. The softmax normalizes
these responses across patch locations.
$\bar f$ denotes the fused visual patch features, and
$q_g$ is derived from the global image feature $g$.
The mappings $\phi_k$ and $\phi_v$ construct attention
keys and values, respectively, incorporating normalization
and channel gating. $\operatorname{MHA}$ denotes multi-head
attention.

We insert the visual condition tokens $\Pi^v$ before
the end-of-text token in the authentic and manipulated
class prompts, each containing learnable class-specific
context tokens. A frozen text encoder with learnable
deep prompts produces the corresponding image-conditioned
representations $t_0$ and $t_1$.

Given the image label $y\in\{0,1\}$, we supervise the
similarities between the global image feature $g$ and
these two representations using
\begin{equation}
\mathcal{L}_{\mathrm{text}}
=
-\log
\frac{
\exp(\widehat{g}^{\top}\widehat{t}_y/\tau_t)
}{
\exp(\widehat{g}^{\top}\widehat{t}_0/\tau_t)
+
\exp(\widehat{g}^{\top}\widehat{t}_1/\tau_t)
}
\label{eq:method_text}
\end{equation}
Here, $y=0$ denotes an authentic image and $y=1$ a
manipulated image, a hat denotes $\ell_2$ normalization,
and $\tau_t$ is a temperature parameter.
This objective jointly updates the prompt-conditioning
branch and the shared visual adaptation parameters.

\paragraph{Hierarchical trace contrastive learning.}
The image--text objective aligns global image features
with image-conditioned class representations. To extend
this supervision to local features, HTCL uses these
representations to select patches for contrastive
supervision at multiple visual layers.

For an image with label $y\in\{0,1\}$, we compute
the compatibility between each patch and the corresponding
text representation at visual layer $\ell$:
\begin{equation}
s_{jy}^{(\ell)}
=
\frac{
\exp\!\left(
(\widehat{f}_j^{(\ell)})^\top t_y/\tau_h
\right)
}{
\exp\!\left(
(\widehat{f}_j^{(\ell)})^\top t_0/\tau_h
\right)
+
\exp\!\left(
(\widehat{f}_j^{(\ell)})^\top t_1/\tau_h
\right)
},
\label{eq:method_candidate_scores}
\end{equation}
where $\widehat{f}_j^{(\ell)}$ is the normalized patch
representation at location $j$ in the image--text
embedding space.
$t_0$ and $t_1$ are the unnormalized image-conditioned
text representations for the authentic and manipulated
classes, respectively, and $\tau_h$ is a temperature
parameter.

We use $s_{jy}^{(\ell)}$ for both candidate filtering
and ranking:
\begin{equation}
\begin{aligned}
\Omega_y^{(\ell)}
&=
\left\{
j \mid e_j>\eta_e,\;
s_{jy}^{(\ell)}>\eta_s
\right\},\\
\mathcal{P}_y^{(\ell)}
&=
\operatorname{TopK}_{B_p}
\left(
\Omega_y^{(\ell)};\,
s_{\cdot y}^{(\ell)}
\right),
\end{aligned}
\label{eq:method_patch_selection}
\end{equation}
where $\eta_e$ and $\eta_s$ are fixed thresholds for
the edge response and patch--text score, respectively.
$\operatorname{TopK}_{B_p}$ returns the indices of
at most $B_p$ candidates in $\Omega_y^{(\ell)}$
with the highest scores $s_{jy}^{(\ell)}$.
Each selected patch receives the image label $y$
as its weak class label.

At layer $\ell$, we project and normalize the features
of patches indexed by $\mathcal{P}_y^{(\ell)}$.
The resulting queries with weak label $y$ form
$\mathcal{Q}_y^{(\ell)}$ within the current minibatch
on each device. We denote each query by $q_r^{(\ell)}$,
where $r$ indexes selected patches across images.

For each visual layer $\ell$ and weak class $y$, we maintain
a memory queue $\mathcal{K}_y^{(\ell)}$ of normalized
historical patch keys.
Since patches sharing a weak label may exhibit different
local trace patterns, each query is paired with its most
similar same-class key, as defined in
Eq.~\eqref{eq:method_positive_reference}:
\begin{equation}
k_{r,+}^{(\ell)}
=
\operatorname*{arg\,max}_{k\in\mathcal{K}_y^{(\ell)}}
(q_r^{(\ell)})^\top k.
\label{eq:method_positive_reference}
\end{equation}
The dot product measures cosine similarity because both
queries and keys are normalized.
Using the selected reference, we compute the contrastive
loss in Eq.~\eqref{eq:method_htcl}:
\begin{equation}
\mathcal{L}_r^{(\ell)}
=
-\log
\frac{
\exp\!\left(
(q_r^{(\ell)})^\top k_{r,+}^{(\ell)}/\tau_e
\right)
}{
\begin{aligned}
&\sum_{q_s^{(\ell)}\in\mathcal{Q}_y^{(\ell)}}
\exp\!\left(
(q_r^{(\ell)})^\top k_{s,+}^{(\ell)}/\tau_e
\right)\\[-2pt]
&\quad+
\sum_{k\in\mathcal{K}_{1-y}^{(\ell)}}
\exp\!\left(
(q_r^{(\ell)})^\top k/\tau_e
\right)
\end{aligned}
},
\label{eq:method_htcl}
\end{equation}
where $\tau_e$ is a temperature parameter.
The first denominator term contains one reference per
current same-class query, including the query's own
positive reference. References selected by other queries
contribute competing terms, with repeated selections of
the same key counted separately.
The second term contains all opposite-class memory keys.

We average the query losses within each weak class,
sum the class objectives, and average across the
selected visual layers:
\begin{equation}
\mathcal{L}_{\mathrm{HTCL}}
=
\frac{1}{|\mathcal{S}|}
\sum_{\ell\in\mathcal{S}}
\sum_{\substack{
y\in\{0,1\}\\
|\mathcal{Q}_y^{(\ell)}|>0
}}
\frac{1}{|\mathcal{Q}_y^{(\ell)}|}
\sum_{q_r^{(\ell)}\in\mathcal{Q}_y^{(\ell)}}
\mathcal{L}_r^{(\ell)},
\label{eq:method_htcl_total}
\end{equation}
where $\mathcal{S}$ denotes the set of selected visual
layers, and each nonempty query set
$\mathcal{Q}_y^{(\ell)}$ contributes a class-wise
mean loss.

\subsection{Relation-Aware Trace Discrimination}
\label{sec:relation_discrimination}

TOVA learns local manipulation representations.
RATD uses local context and intra-image references to
distinguish authentic and manipulated regions with
similar appearance. We first construct local evidence
from contextual, structural, and original-image detail
features, and then incorporate regional comparison
messages into localization.

\paragraph{Local Evidence Construction.}
To characterize local variation relative to the surrounding
content, we combine each patch feature with center--surround
statistics at two scales, as defined in
Eq.~\eqref{eq:method_context}:
\begin{equation}
\begin{aligned}
\xi_j^{(s)}
&=
\mu_j^{(r_s)}
\Vert \nu_j^{(r_s,R_s)}
\Vert
\big|\mu_j^{(r_s)}-\nu_j^{(r_s,R_s)}\big|,\\
C_j
&=
\Phi_C\!\left(
\bar f_j\Vert\xi_j^{(1)}\Vert\xi_j^{(2)}
\right),
\end{aligned}
\label{eq:method_context}
\end{equation}
where $\bar f_j$ is the fused visual patch feature from
Sec.~\ref{sec:trace_adaptation}, and $\Vert$ denotes
channel concatenation.
For each scale $s\in\{1,2\}$, $\mu_j^{(r_s)}$ is the
mean feature within a central window of width $r_s$,
and $\nu_j^{(r_s,R_s)}$ is the mean over the surrounding
ring between widths $r_s$ and $R_s$.
The learnable mapping $\Phi_C$ produces the contextual
feature $C_j$ from these concatenated features.

We complement the patch-grid context with spatial structure
from the resized image and local details from the original
image. Since resizing can attenuate fine local variations,
the detail branch retains the original RGB image and its
local mean residual. The branch computations and feature
fusion are defined in Eq.~\eqref{eq:method_native_details}:
\begin{equation}
\begin{aligned}
D
&=\Phi_D\!\left(
\bar f,\Phi_{\mathrm{RGB}}(I_{\mathrm{res}})
\right),\\
T
&=\Phi_T\!\left(
I\Vert
\big(I-\operatorname{Mean}_{3\times3}(I)\big)
\right),\\
F_i
&=C_i\Vert D_i\Vert T_i,
\end{aligned}
\label{eq:method_native_details}
\end{equation}
where $I$ is the original image and $I_{\mathrm{res}}$
is its resized version.
The structural branch $\Phi_D$ upsamples the visual
features and fuses them with shallow RGB features
extracted by $\Phi_{\mathrm{RGB}}$.
The detail branch $\Phi_T$ encodes the original image
and its residual from a fixed $3\times3$ averaging filter,
which captures local intensity variations.

Before forming $F_i$, we align the three branch outputs
on a common localization grid expressed in original-image
coordinates. We use $i$ to index this aligned grid and
$j$ to index the visual patch grid.

\paragraph{Intra-Image Reference Comparison.}
Local features encode both image content and processing
traces. Regions with similar appearance therefore provide
references for comparing their local evidence.
We establish intra-image appearance correspondences
using SIFT~\cite{lowe2004distinctive} on the original image,
retaining spatially separated mutual matches with locally
consistent displacements.
SIFT identifies the regions to compare, while descriptors
pooled from $F$ represent their learned local evidence.

We map the matched keypoints to the localization grid
and construct a directed correspondence set $\mathcal M$
containing both directions of each retained match.
For $(u,v)\in\mathcal M$, $u$ is the receiving location
and $v$ is its reference location.
Each correspondence has a confidence $w_{uv}$ determined
by the SIFT descriptor distance and support from nearby
matches with consistent displacements.

For each directed correspondence, we pool local descriptors
and construct a comparison message using
Eq.~\eqref{eq:method_messages}:
\begin{equation}
\begin{aligned}
d_z
&=
\frac{
\sum_{i\in\mathcal N(z)}a_iF_i
}{
\sum_{i\in\mathcal N(z)}a_i
},
\quad z\in\{u,v\},\\
\mathbf b_{u\leftarrow v}
&=
\psi\!\left(
d_v\Vert(d_v-d_u)\Vert w_{uv}
\right),
\end{aligned}
\label{eq:method_messages}
\end{equation}
where $\mathcal N(z)$ is a neighborhood of $z$ on the
localization grid, and $a_i$ is the valid original-image
area represented by grid location $i$.
The shared MLP $\psi$ encodes the reference descriptor,
its signed difference from the receiving descriptor,
and the correspondence confidence.

We incorporate the sparse comparisons into dense prediction
by aggregating messages around their receiving locations.
Let $\mathcal M_i$ contain directed correspondences whose
receiving locations lie within an aggregation neighborhood
of $i$. The confidence-weighted aggregation and localization
prediction are defined in
Eq.~\eqref{eq:method_message_aggregation}:
\begin{equation}
\begin{aligned}
\kappa_i
&=
\sum_{(z,z')\in\mathcal M_i}w_{zz'},\\
P_i
&=
\frac{
\sum_{(z,z')\in\mathcal M_i}
w_{zz'}\mathbf b_{z\leftarrow z'}
}{
\max(1,\kappa_i)
},\\
m
&=
H(F\Vert P\Vert\kappa),
\end{aligned}
\label{eq:method_message_aggregation}
\end{equation}
where $\kappa_i$ records the accumulated correspondence
confidence. The convolutional head $H$ combines local
features, comparison messages, and confidence to predict
spatial manipulation logits $m$.

We supervise the spatial predictions using only image-level
labels through area-weighted multiple instance learning (MIL).
We first apply local area-weighted smoothing to $m$,
yielding $\bar m$, and then aggregate the highest-scoring
regions using the objective in Eq.~\eqref{eq:method_mil}:
\begin{equation}
\mathcal L_{\mathrm{MIL}}(I,y)
=
\operatorname{BCE}\!\left(
\sigma\!\left(
\operatorname{TopArea}_{\rho}(\bar m)
\right),y
\right),
\label{eq:method_mil}
\end{equation}
where $\sigma$ denotes the sigmoid function and $y$
is the image-level authenticity label.
$\operatorname{TopArea}_{\rho}$ computes the area-weighted
mean of the highest-scoring logits covering a fraction
$\rho$ of the valid image area.

We combine MIL with the image--text objective in
Eq.~\eqref{eq:method_text} and HTCL in
Eq.~\eqref{eq:method_htcl_total}.
The full training objective is given in
Eq.~\eqref{eq:method_objective}:
\begin{equation}
\mathcal L
=
\mathcal L_{\mathrm{MIL}}
+
\mathcal L_{\mathrm{text}}
+
\lambda_e(t)\mathcal L_{\mathrm{HTCL}},
\label{eq:method_objective}
\end{equation}
where $\lambda_e(t)$ controls the contrastive weight
at training step $t$.
MIL supervises the spatial predictions, while the
image--text and contrastive objectives adapt the shared
visual representations at the image and patch levels,
respectively.

%% file: sec/4_experiment.tex
\section{Experiments}
\label{sec:experiments}

\begin{table*}[ht]
\centering
\setlength{\belowcaptionskip}{3pt}
\caption{Comparison of image manipulation detection and localization methods.
I-F1 and P-F1 denote image-level and pixel-level F1, respectively.
Average I-F1 and P-F1 are computed from the listed scores over four
and five datasets, respectively. NIST16 is evaluated for localization only.
\textbf{Bold} and \underline{underlined} values indicate the best and
second-best weakly supervised results, respectively.
Available baseline I-F1 scores, together with P-F1 scores for WSCCL
and ViLaCo, follow Sheng et al.~\cite{sheng2026weakly}.
Dashes indicate unavailable scores. Un., Full., and Weak. denote
unsupervised, fully supervised, and weakly supervised methods.}
\label{tab:state}
\adjustbox{max width=0.9\textwidth}{%
\begin{tabular}{l|l|c|cc|cc|cc|cc|c|cc}
\toprule
 & \multirow{2}{*}{\textbf{Method}} & \multirow{2}{*}{\textbf{Source}}
 & \multicolumn{2}{c|}{\textbf{CASIAv1}}
 & \multicolumn{2}{c|}{\textbf{Columbia}}
 & \multicolumn{2}{c|}{\textbf{COVERAGE}}
 & \multicolumn{2}{c|}{\textbf{IMD2020}}
 & \multicolumn{1}{c|}{\textbf{NIST16}}
 & \multicolumn{2}{c}{\textbf{Average}} \\
 & & & I-F1 & P-F1 & I-F1 & P-F1 & I-F1 & P-F1
 & I-F1 & P-F1 & P-F1 & I-F1 & P-F1 \\
\hline
\multirow{2}{*}{\rotatebox{90}{\textbf{Un.}}} & NOI1\cite{mahdian2009using} & \textit{\small IVC09}
  & 0.000 & 0.157 & 0.000 & 0.311 & 0.000 & 0.205 & 0.000 & 0.124 & 0.089 & 0.000 & 0.177 \\
 & CFA1\cite{ferrara2012image} & \textit{\small TIFS12}
  & 0.000 & 0.140 & 0.000 & 0.320 & 0.000 & 0.188 & 0.000 & 0.111 & 0.106 & 0.000 & 0.173 \\
\hline
\multirow{6}{*}{\rotatebox{90}{\textbf{Full.}}} & ManTra-Net\cite{wu2019mantra} & \textit{\small CVPR19}
  & 0.000 & 0.155 & 0.000 & 0.364 & 0.000 & 0.286 & 0.000 & 0.122 & 0.000 & 0.000 & 0.185 \\
 & CAT-Net\cite{kwon2021cat} & \textit{\small WACV21}
  & 0.459 & 0.276 & 0.505 & 0.352 & 0.169 & 0.134 & 0.229 & 0.102 & 0.138 & 0.341 & 0.200 \\
 & MVSS-Net\cite{chen2021image} & \textit{\small ICCV21}
  & 0.758 & 0.452 & 0.802 & 0.638 & 0.244 & 0.453 & 0.355 & 0.260 & 0.292 & 0.540 & 0.419 \\
 & GSR-Net\cite{zhou2020generate} & \textit{\small AAAI20}
  & 0.022 & 0.387 & 0.022 & 0.613 & 0.000 & 0.285 & 0.014 & 0.175 & 0.283 & 0.015 & 0.349 \\
 & RRU-Net\cite{bi2019rru} & \textit{\small CVPRW19}
  & 0.001 & 0.225 & 0.000 & 0.452 & 0.000 & 0.189 & 0.000 & 0.232 & 0.265 & 0.000 & 0.273 \\
 & FCN+DA\cite{chen2021image} & \textit{\small ICCV21}
  & 0.775 & 0.441 & 0.481 & 0.223 & 0.180 & 0.199 & 0.182 & 0.270 & 0.167 & 0.405 & 0.260 \\
\hline
\multirow{9}{*}{\rotatebox{90}{\textbf{Weak.}}} & MIL-FCN\cite{pathak2014fully} & \textit{\small ICLRW15}
  & 0.553 & 0.117 & 0.338 & 0.089 & 0.115 & 0.121 & 0.205 & 0.097 & 0.024 & 0.303 & 0.090 \\
 & MIL-FCN+WSCL\cite{zhai2023towards} & \textit{\small ICCV23}
  & 0.738 & 0.172 & 0.680 & 0.270 & 0.544 & 0.178 & 0.360 & 0.193 & 0.110 & 0.581 & 0.185 \\
 & Araslanov\cite{araslanov2020single} & \textit{\small CVPR20}
  & 0.496 & 0.112 & 0.140 & 0.102 & 0.140 & 0.127 & 0.219 & 0.094 & 0.026 & 0.249 & 0.092 \\
 & Araslanov+WSCL\cite{zhai2023towards} & \textit{\small ICCV23}
  & 0.679 & 0.153 & 0.483 & 0.362 & 0.348 & 0.201 & 0.316 & 0.173 & 0.099 & 0.457 & 0.198 \\
 & M$^2$RL-Net\cite{li2025m2rl} & \textit{\small AAAI25}
  & \underline{0.866} & 0.347 & \textbf{0.975} & 0.434 & 0.610 & 0.213 & 0.585 & 0.248 & 0.113 & 0.759 & 0.271 \\
 & SO-WSL\cite{zhu2025self} & \textit{\small ICASSP25}
  & -- & 0.334 & -- & 0.385 & -- & 0.239 & -- & 0.259 & \textbf{0.288} & -- & 0.301 \\
 & WSCCL\cite{bai2025weakly} & \textit{\small KBS25}
  & -- & 0.347 & -- & 0.273 & -- & 0.301 & -- & 0.265 & 0.159 & -- & 0.269 \\
 & ViLaCo\cite{sheng2026weakly} & \textit{\small AAAI26}
  & \textbf{0.880} & \textbf{0.491} & \underline{0.917} & \underline{0.536} & \underline{0.630} & \underline{0.319} & \underline{0.675} & \underline{0.365} & 0.267 & \underline{0.776} & \underline{0.396} \\
 & \textbf{Ours} & -
  & 0.837 & \underline{0.463} & 0.830 & \textbf{0.598} & \textbf{0.719} & \textbf{0.323} & \textbf{0.935} & \textbf{0.410} & \underline{0.278} & \textbf{0.830} & \textbf{0.414} \\
\bottomrule
\end{tabular}%
}
\end{table*}

\begin{table}[t]
\centering
\caption{Ablation of TOVA and RATD.
TOVA comprises Trace-Oriented Prompt Conditioning and HTCL;
RATD comprises Local Evidence Construction and Intra-Image Reference Comparison.
LoRA remains enabled in all variants. A dash denotes a disabled component.
Avg.\ denotes the mean over CASIAv1 and IMD2020.
The best Avg.\ scores are in \textbf{bold}.}
\vspace{-0.3cm}  % 减少表格和正文之间的空
\label{tab:ablation_components}
\setlength{\belowcaptionskip}{-0.1cm}
\resizebox{\linewidth}{!}{%
\begin{tabular}{cc|cc|cc|c|c}
\toprule
\multirow{2}{*}{\textbf{TOVA}} & \multirow{2}{*}{\textbf{RATD}}
& \multicolumn{2}{c|}{\textbf{CASIAv1}}
& \multicolumn{2}{c|}{\textbf{IMD2020}}
& \multicolumn{2}{c}{\textbf{Avg.}} \\
&& I-F1 & P-F1 & I-F1 & P-F1 & I-F1 & P-F1 \\
\hline
- & - & 0.712 & 0.243 & 0.823 & 0.287 & 0.768 & 0.265 \\
$\checkmark$ & - & 0.771 & 0.350 & 0.901 & 0.358 & 0.836 & 0.354 \\
- & $\checkmark$ & 0.758 & 0.373 & 0.872 & 0.324 & 0.815 & 0.349 \\
$\checkmark$ & $\checkmark$ & 0.837 & 0.463 & 0.935 & 0.410 & \textbf{0.886} & \textbf{0.437} \\
\bottomrule
\end{tabular}%
}
\vspace{-0.5cm}  % 减少表格和正文之间的空白
\end{table}

\subsection{Implementation Details}
\label{sec:implementation}

Following prior weakly supervised methods~\cite{zhai2023towards,li2025m2rl,sheng2026weakly},
we train on CASIAv2~\cite{dong2013casia} using only image-level
authenticity labels. We evaluate in-domain performance on CASIAv1
and cross-dataset generalization on Columbia~\cite{hsu2006detecting},
COVERAGE~\cite{wen2016coverage}, IMD2020~\cite{novozamsky2020imd2020},
and NIST16~\cite{nist2016nimble}. We report image-level F1 (I-F1)
for detection and pixel-level F1 (P-F1) for localization;
NIST16 is used only for localization.

We use the pretrained CLIP~\cite{radford2021learning}
ViT-L/14@336px backbone. Images are resized to a longer side
of 512 pixels while preserving their aspect ratio, then centrally
padded to $512\times512$ for visual encoding. Original-resolution
images are retained for detail extraction and intra-image
correspondence construction. We jointly optimize the visual
LoRA adapters~\cite{hu2021lora}, associated attention biases,
learnable prompts, and task-specific modules, while keeping
the remaining pretrained CLIP parameters frozen.

Our framework is implemented in PyTorch 2.0 and trained on
eight NVIDIA A100 GPUs. We use
AdamW~\cite{loshchilov2019decoupledweightdecayregularization}
with a batch size of 12 and a learning rate of $10^{-5}$
for 30 epochs. The learnable prompt length is 12.
We extract visual features from layers 6, 12, 18, and 24
and maintain an HTCL memory queue of 1,024 keys for each
class at each selected layer. The contrastive weight
$\lambda_e(t)$ is linearly increased to 0.1 during training.

\subsection{Comprehensive Evaluation}

\paragraph{State-of-the-Art Comparison.}
Table~\ref{tab:state} evaluates the framework for both image-level
detection and pixel-level localization. Our method achieves
average I-F1 and P-F1 scores of 0.830 and 0.414, respectively,
the highest among the evaluated weakly supervised methods.
It obtains the best I-F1 on COVERAGE and IMD2020, reaching
0.719 and 0.935, and the highest weakly supervised P-F1 on
Columbia, COVERAGE, and IMD2020. Since these datasets are
not used for training, the localization results demonstrate
that the framework generalizes beyond its CASIAv2 training
data. The detection and localization results together
show that image-level training supports spatial predictions
alongside global authenticity decisions. This supports
the overall design of our framework: TOVA connects
authenticity learning to local patch representations,
while RATD uses contextual evidence and intra-image
comparisons to interpret these representations for localization.

\begin{figure}[t]
\centering
\includegraphics[width=0.48\textwidth]{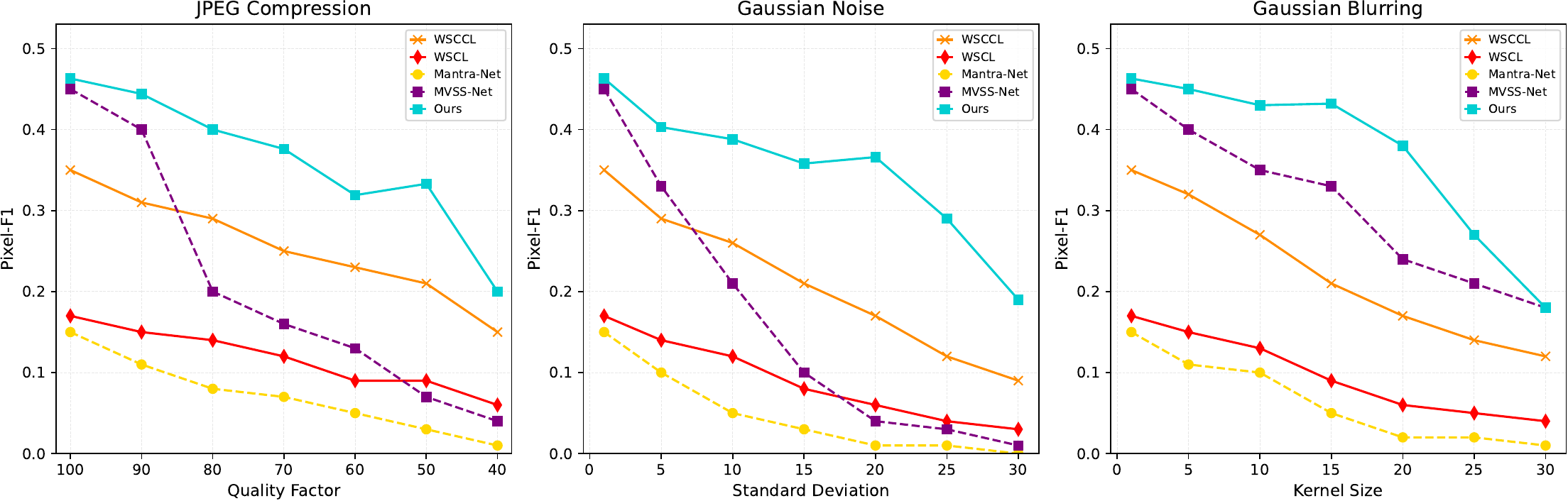}
\caption{Localization robustness on CASIAv1 under JPEG compression
(left), Gaussian noise (middle), and Gaussian blurring (right),
evaluated using P-F1. Dashed and solid lines denote fully
supervised and weakly supervised methods, respectively.}
\label{fig:robustness}
\vspace{-0.5cm}  % 减少表格和正文之间的空白
\end{figure}

\paragraph{Robustness to Image Perturbations.}
\label{sec:robustness}
We evaluate localization robustness on CASIAv1 under JPEG
compression, Gaussian noise, and Gaussian blurring.
These perturbations alter local image statistics and details,
testing whether the learned manipulation evidence remains
useful after post-processing. As shown in Fig.~\ref{fig:robustness},
our method achieves the highest or joint-highest P-F1 across
all tested settings. At a JPEG quality factor of 40, a noise
standard deviation of 30, and a blur kernel size of 30,
P-F1 reaches 0.20, 0.19, and 0.18, respectively. Under
moderate blurring, P-F1 remains above 0.42 for kernel sizes
up to 15. The framework thus retains useful regional
predictions when post-processing changes the local cues
used for localization, extending the evaluation of its
jointly learned representations and regional discrimination
to perturbed inputs.

\begin{figure}[t]
\centering
\includegraphics[width=0.38\textwidth]{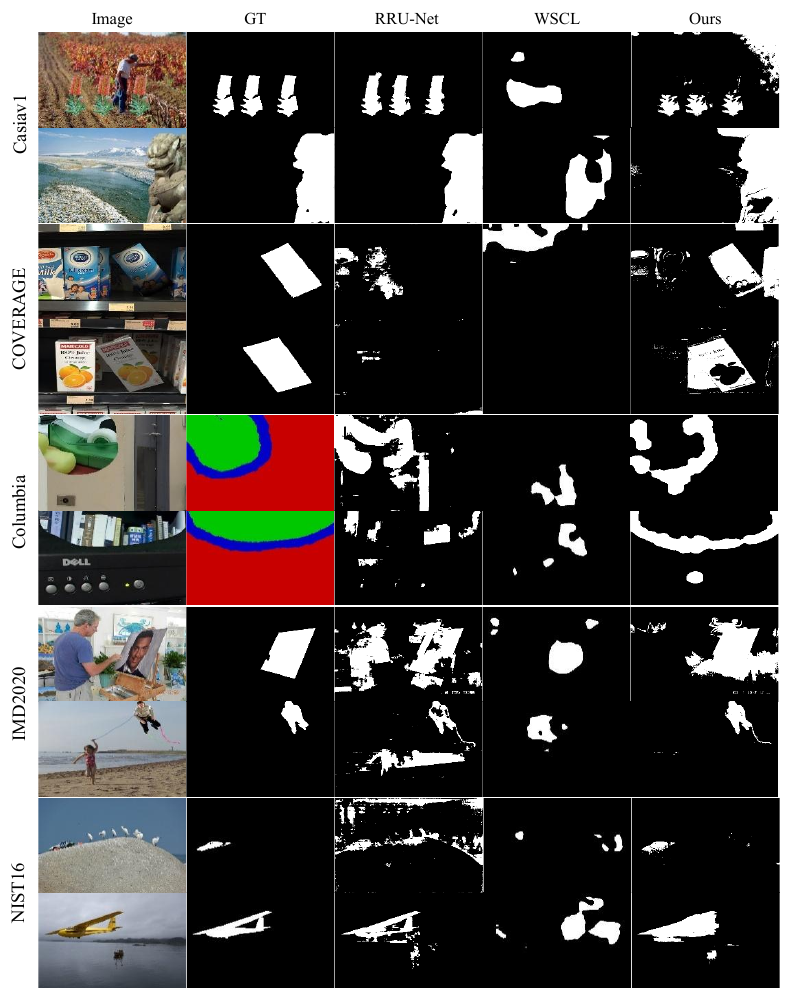}
\caption{Qualitative localization results on CASIAv1,
COVERAGE, Columbia, IMD2020, and NIST16.
Columns show the input image, ground-truth annotations
(GT), and predictions from RRU-Net, WSCL, and our method.
White regions in the prediction maps indicate pixels
predicted as manipulated.}
\label{fig:qualitative}
\vspace{-0.5cm}  % 减少表格和正文之间的空白
\end{figure}

\paragraph{Qualitative Results.}
Fig.~\ref{fig:qualitative} presents localization examples
from five datasets. In COVERAGE, our method identifies
the manipulated product packages within shelves containing
objects with similar appearance. Such scenes illustrate
the motivation for RATD: local authenticity evidence needs
to be interpreted in relation to surrounding content and
appearance-related regions. The IMD2020 predictions locate
the edited portrait and person regions, while the NIST16
examples show responses at both a small edited region and
the aircraft. These cases complement the quantitative
results by showing the spatial correspondence between
predictions and target manipulations across different
contents and sizes. They illustrate the regional localization
behavior sought by combining trace-sensitive representations
with contextual and intra-image reference information.

\subsection{Ablation Study}

We conduct ablation studies on CASIAv1 and IMD2020,
reporting detection and localization performance on each
dataset and their averages.

\paragraph{Overall Components.}
Table~\ref{tab:ablation_components} evaluates the contributions
of trace representation learning and regional discrimination,
with LoRA enabled in every configuration. TOVA and RATD
individually raise average P-F1 from 0.265 to 0.354 and 0.349,
respectively. Their combination achieves an average I-F1
of 0.886 and P-F1 of 0.437, improving localization by
17.2 percentage points over the baseline. TOVA adapts the
visual features through image-conditioned class representations
and local contrastive supervision, while RATD uses contextual
evidence and regional comparisons to interpret these features
for dense prediction. The additional P-F1 gains of 8.3 and
8.8 points over the individual components support coupling
these two stages: learning local manipulation representations
and using regional relations to guide their spatial predictions.

\paragraph{TOVA Components.}
With RATD enabled, Table~\ref{tab:ablation_tova} examines how
Trace-Oriented Prompt Conditioning and HTCL contribute to
local representation learning. Prompt conditioning raises
average P-F1 from 0.349 to 0.388, while learnable class
prompts are retained in both configurations. This gain
supports conditioning class representations on image content
and local structural cues. HTCL alone achieves 0.393,
demonstrating the benefit of direct contrastive constraints
on selected patches. Combining both components further
increases average P-F1 to 0.437, a gain of 4.4 percentage
points over HTCL alone. The conditioned class representations
participate in image--text alignment and guide patch selection
for HTCL. The joint improvement therefore supports connecting
prompt adaptation with patch-level supervision, enabling
image-level authenticity learning to shape the local visual
features used for localization.

\begin{table}[t]
\centering
\caption{Ablation of the TOVA components with RATD enabled.
Prompt denotes Trace-Oriented Prompt Conditioning.
Learnable class prompts are retained when conditioning is disabled.
Avg.\ denotes the mean over CASIAv1 and IMD2020.
The best Avg.\ scores are in \textbf{bold}.}
\vspace{-0.3cm}  % 减少表格和正文之间的空白
\label{tab:ablation_tova}
\setlength{\belowcaptionskip}{-0.1cm}
\resizebox{0.9\linewidth}{!}{%
\begin{tabular}{cc|cc|cc|c|c}
\toprule
\multirow{2}{*}{\textbf{Prompt}} & \multirow{2}{*}{\textbf{HTCL}}
& \multicolumn{2}{c|}{\textbf{CASIAv1}}
& \multicolumn{2}{c|}{\textbf{IMD2020}}
& \multicolumn{2}{c}{\textbf{Avg.}} \\
&& I-F1 & P-F1 & I-F1 & P-F1 & I-F1 & P-F1 \\
\hline
- & - & 0.758 & 0.373 & 0.872 & 0.324 & 0.815 & 0.349 \\
$\checkmark$ & - & 0.784 & 0.426 & 0.893 & 0.350 & 0.839 & 0.388 \\
- & $\checkmark$ & 0.792 & 0.421 & 0.923 & 0.365 & 0.858 & 0.393 \\
$\checkmark$ & $\checkmark$ & 0.837 & 0.463 & 0.935 & 0.410 & \textbf{0.886} & \textbf{0.437} \\
\bottomrule
\end{tabular}%
}
\vspace{-0.3cm}  % 减少表格和正文之间的空白
\end{table}

\paragraph{RATD Components.}
Table~\ref{tab:ablation_ratd} evaluates Local Evidence
Construction and Intra-Image Reference Comparison with TOVA
enabled. Local evidence construction raises average P-F1
from 0.354 to 0.389, demonstrating the benefit of combining
contextual, structural, and original-image detail features.
Reference comparison alone achieves 0.409 using projected
visual features, indicating that comparisons between
appearance-matched regions provide useful information for
localization. Combining both components reaches 0.437,
improving average P-F1 by 2.8 points over reference comparison
alone and 4.8 points over local evidence construction alone.
Thus, richer local evidence benefits localization when
regional comparisons are present, and comparison messages
provide an additional gain when the local evidence branches
are enabled. These results support interpreting manipulation
evidence jointly through its surroundings and corresponding
regions within the image.

\begin{table}[t]
\centering
\caption{Ablation of the RATD components with TOVA enabled.
Local and Reference denote Local Evidence Construction and
Intra-Image Reference Comparison, respectively.
Without Local, projected visual features are used for localization
and reference comparison. Avg.\ denotes the mean over CASIAv1 and IMD2020.
The best Avg.\ scores are in \textbf{bold}.}
\vspace{-0.3cm}  % 减少表格和正文之间的空白
\label{tab:ablation_ratd}
\setlength{\belowcaptionskip}{-0.1cm}
\resizebox{0.9\linewidth}{!}{%
\begin{tabular}{cc|cc|cc|c|c}
\toprule
\multirow{2}{*}{\textbf{Local}} & \multirow{2}{*}{\textbf{Reference}}
& \multicolumn{2}{c|}{\textbf{CASIAv1}}
& \multicolumn{2}{c|}{\textbf{IMD2020}}
& \multicolumn{2}{c}{\textbf{Avg.}} \\
&& I-F1 & P-F1 & I-F1 & P-F1 & I-F1 & P-F1 \\
\hline
- & - & 0.771 & 0.350 & 0.901 & 0.358 & 0.836 & 0.354 \\
$\checkmark$ & - & 0.795 & 0.393 & 0.915 & 0.384 & 0.855 & 0.389 \\
- & $\checkmark$ & 0.810 & 0.416 & 0.908 & 0.401 & 0.859 & 0.409 \\
$\checkmark$ & $\checkmark$ & 0.837 & 0.463 & 0.935 & 0.410 & \textbf{0.886} & \textbf{0.437} \\
\bottomrule
\end{tabular}%
}
\vspace{-0.3cm}  % 减少表格和正文之间的空白
\end{table}

\paragraph{HTCL Supervision Layers.}
Table~\ref{tab:ablation_htcl_layers} varies the layers receiving
HTCL supervision while keeping the feature-extraction layers
and fusion architecture unchanged. Extending supervision
from layer 24 to layers 12 and 24 increases average P-F1
from 0.383 to 0.422. Supervising layers 6, 12, 18, and 24
further raises it to 0.437 and achieves the highest average
I-F1 of 0.886. The 5.4-percentage-point localization gain
over final-layer supervision demonstrates the value of
applying patch-level constraints at multiple visual depths.
Intermediate-layer HTCL provides direct class-guided
contrastive constraints on the patch representations
subsequently fused for localization. The results therefore
support distributing local authenticity supervision across
the visual hierarchy and motivate the four-layer setting
used in our framework.

\begin{table}[t]
\centering
\caption{Effect of the visual layers receiving HTCL supervision.
Feature extraction layers and all other components are fixed;
only the layers used for contrastive supervision vary.
Avg.\ denotes the mean over CASIAv1 and IMD2020.
The best Avg.\ scores are in \textbf{bold}.}
\vspace{-0.3cm}  % 减少表格和正文之间的空白
\label{tab:ablation_htcl_layers}
\setlength{\belowcaptionskip}{-0.1cm}
\resizebox{0.9\linewidth}{!}{%
\begin{tabular}{c|cc|cc|c|c}
\toprule
\multirow{2}{*}{\textbf{HTCL layers}}
& \multicolumn{2}{c|}{\textbf{CASIAv1}}
& \multicolumn{2}{c|}{\textbf{IMD2020}}
& \multicolumn{2}{c}{\textbf{Avg.}} \\
& I-F1 & P-F1 & I-F1 & P-F1 & I-F1 & P-F1 \\
\hline
$\{24\}$ & 0.788 & 0.424 & 0.890 & 0.341 & 0.839 & 0.383 \\
$\{12,24\}$ & 0.809 & 0.430 & 0.926 & 0.413 & 0.868 & 0.422 \\
$\{6,12,18,24\}$ & 0.837 & 0.463 & 0.935 & 0.410 & \textbf{0.886} & \textbf{0.437} \\
\bottomrule
\end{tabular}%
}
\vspace{-0.3cm}  % 减少表格和正文之间的空白
\end{table}

\paragraph{Local Evidence Features.}
Table~\ref{tab:ablation_local_features} examines the information
supplied by the three local evidence branches. Adding
contextual features $C$ to the structural and detail branches
improves average P-F1 by 4.7 percentage points, the largest
gain among the three additions. This result supports
interpreting patch responses through center--surround
statistics, which provide regional evidence beyond the
central feature. Adding structural features $D$ or
original-image detail features $T$ to the other two branches
improves average P-F1 by 1.5 and 2.4 points, respectively.
These gains support complementing patch-grid context with
visual--RGB feature fusion and original-resolution information.
Together, the three branches achieve the highest average
I-F1 and P-F1 of 0.886 and 0.437, providing complementary
local evidence for both regional comparison and dense prediction.

\begin{table}[t]
\centering
\caption{Ablation of local evidence features.
$C$, $D$, and $T$ denote contextual, structural, and original-image
detail features, respectively. All other components remain enabled.
Avg.\ denotes the mean over CASIAv1 and IMD2020.
Best scores are in \textbf{bold}.}
\vspace{-0.3cm}  % 减少表格和正文之间的空白
\label{tab:ablation_local_features}
\setlength{\belowcaptionskip}{-0.1cm}
\resizebox{0.9\linewidth}{!}{%
\begin{tabular}{ccc|cc|cc|c|c}
\toprule
\multirow{2}{*}{$\boldsymbol{C}$} & \multirow{2}{*}{$\boldsymbol{D}$} & \multirow{2}{*}{$\boldsymbol{T}$}
& \multicolumn{2}{c|}{\textbf{CASIAv1}}
& \multicolumn{2}{c|}{\textbf{IMD2020}}
& \multicolumn{2}{c}{\textbf{Avg.}} \\
&&& I-F1 & P-F1 & I-F1 & P-F1 & I-F1 & P-F1 \\
\hline
- & $\checkmark$ & $\checkmark$ & 0.802 & 0.397 & 0.910 & 0.382 & 0.856 & 0.390 \\
$\checkmark$ & - & $\checkmark$ & 0.819 & 0.431 & 0.926 & 0.412 & 0.873 & 0.422 \\
$\checkmark$ & $\checkmark$ & - & 0.811 & 0.422 & 0.930 & 0.404 & 0.871 & 0.413 \\
$\checkmark$ & $\checkmark$ & $\checkmark$ & 0.837 & 0.463 & 0.935 & 0.410 & \textbf{0.886} & \textbf{0.437}  \\
\bottomrule
\end{tabular}%
}
\vspace{-0.5cm}  % 减少表格和正文之间的空白
\end{table}

%% file: sec/5_conclusion.tex
\section{Conclusion}
\label{sec:conclusion}

We presented a CLIP-based framework combining Trace-Oriented
Vision--Language Adaptation and Relation-Aware Trace Discrimination
for weakly supervised image manipulation localization.
Experiments demonstrate strong detection and localization
performance and robustness, with ablations supporting
the components' complementary roles.

%% file: sec/6_LLM.tex
\section*{LLM Usage Statement}
We used OpenAI's GPT models through ChatGPT and Codex
to assist with manuscript revision and code development.
This assistance included language polishing, improving
text organization, checking mathematical notation,
discussing methodological descriptions, preparing
related-work summaries, and supporting code implementation
and debugging. The authors take full responsibility for
the scientific content, experimental results, code,
and final manuscript.